\documentclass[letterpaper, 10 pt, conference]{ieeeconf}  

\IEEEoverridecommandlockouts                              

\usepackage{graphicx}
\usepackage{lipsum}  
\usepackage{xcolor}
\usepackage{tabularx}
\usepackage{hhline}
\usepackage{multirow}
\usepackage{algorithm}
\usepackage{algorithmic}
\usepackage{subcaption}
\usepackage{amsmath}
\usepackage{amsfonts}
\usepackage{siunitx}

\title{\LARGE \bf
PUCK: Parallel Surface and Convolution-kernel Tracking for Event-Based Cameras
}

\author{{Luna Gava$^{1,2}$, Marco Monforte$^{2}$, Massimiliano Iacono$^{2}$,  Chiara Bartolozzi$^{2}$, Arren Glover$^{2}$}
\thanks{$^{1}$ Luna Gava is with the Department of Computer Science, The University of Manchester, Manchester, UK. \{\tt\small luna.gava@postgrad.manchester.ac.uk}
\thanks{$^{2}$ All the authors are with the Event-Driven Perception for Robotics Research Line, Istituto Italiano di Tecnologia, Italy. \{\tt\small \{marco.monforte, massimiliano.iacono, arren.glover, chiara.bartolozzi\}@iit.it}}

\begin{document}

\maketitle
\thispagestyle{empty}
\pagestyle{empty}

\begin{abstract}
Low latency and accuracy are fundamental requirements when vision is integrated in robots for high-speed interaction with targets, since they affect system reliability and stability. In such a scenario, the choice of the sensor and algorithms is important for the entire control loop. The technology of event-cameras can guarantee fast visual sensing in dynamic environments, but requires a tracking algorithm that can keep up with the high data rate induced by the robot ego-motion while maintaining accuracy and robustness to distractors. 
In this paper, we introduce a novel tracking method that leverages the Exponential Reduced Ordinal Surface (EROS) data representation to decouple event-by-event processing and tracking computation. The latter is performed using convolution kernels to detect and follow a circular target moving on a plane.
To benchmark state-of-the-art event-based tracking, we propose the task of tracking the air hockey puck sliding on a surface, with the future aim of controlling the iCub robot to reach the target precisely and on time. Experimental results demonstrate that our algorithm achieves the best compromise between low latency and tracking accuracy both when the robot is still and when moving.
\end{abstract}


\section{INTRODUCTION}

Autonomous robots working in highly dynamic and unconstrained environments can benefit of high-speed, low-latency vision integrated on the robot.
Some robotic applications suffice with ``look-before-moving'', whereby the action is performed in open-loop; however, in dynamic environments, continuous feedback and corrections are required. In such a ``look-while-moving'' paradigm, the visual signal depends also on the robot's motion. Vision in the control loop raises issues relating to the continuous data rate combined with low latency requirements. The latter causes a visual feedback delay that could affect the system stability, even though the robot is capable of moving very quickly. Selection of appropriate sensors and algorithms is therefore crucial.

To explore the role of visual sensing in tracking for high-speed robotics and the challenges behind it, we propose the game of air-hockey.
The air hockey game is a 2D constrained environment, characterized by a high-velocity puck (e.g., 15\,m/s in professional games), fast changes of direction, and a short motion range. This means that for a 2\,m long table, the robot has at most 130\,ms (similar to the human vision reflex~\cite{thorpe1996speed}) to find the target and react. Solving air hockey with a robot has been achieved using both single~\cite{alattar2019autonomous} and multiple~\cite{liu2021efficient, namiki2013hierarchical} cameras, highlighting that high frame rate sensors (i.e., 120-500\,fps) are required to accurately track the puck. The performance benefits from the use of high-speed manipulators, such as KUKA~\cite{liu2021efficient} or Franka Emika Panda arm~\cite{alattar2019autonomous} or ad-hoc robots specifically built for the task~\cite{namiki2013hierarchical, bentivegna2004learning}. However, it has also been shown that a humanoid robot can play air hockey~\cite{bentivegna2004learning} using its own eyes (cameras) and learning to play from observing a human player and from practice. To solve the problem most conveniently, most of the works used fixed cameras mounted above the table at a height sufficient to see the whole playing field. This approach, while solving the specific problem, does not scale to autonomous robots, that need to mount the vision sensors on the platform so that the robot can perform other dynamic tasks in different environments. Solving the specific air-hockey task could be a good starting point to study the problem of perception when coupled with action.
\begin{figure}
    \centering
    \includegraphics[width=\linewidth]{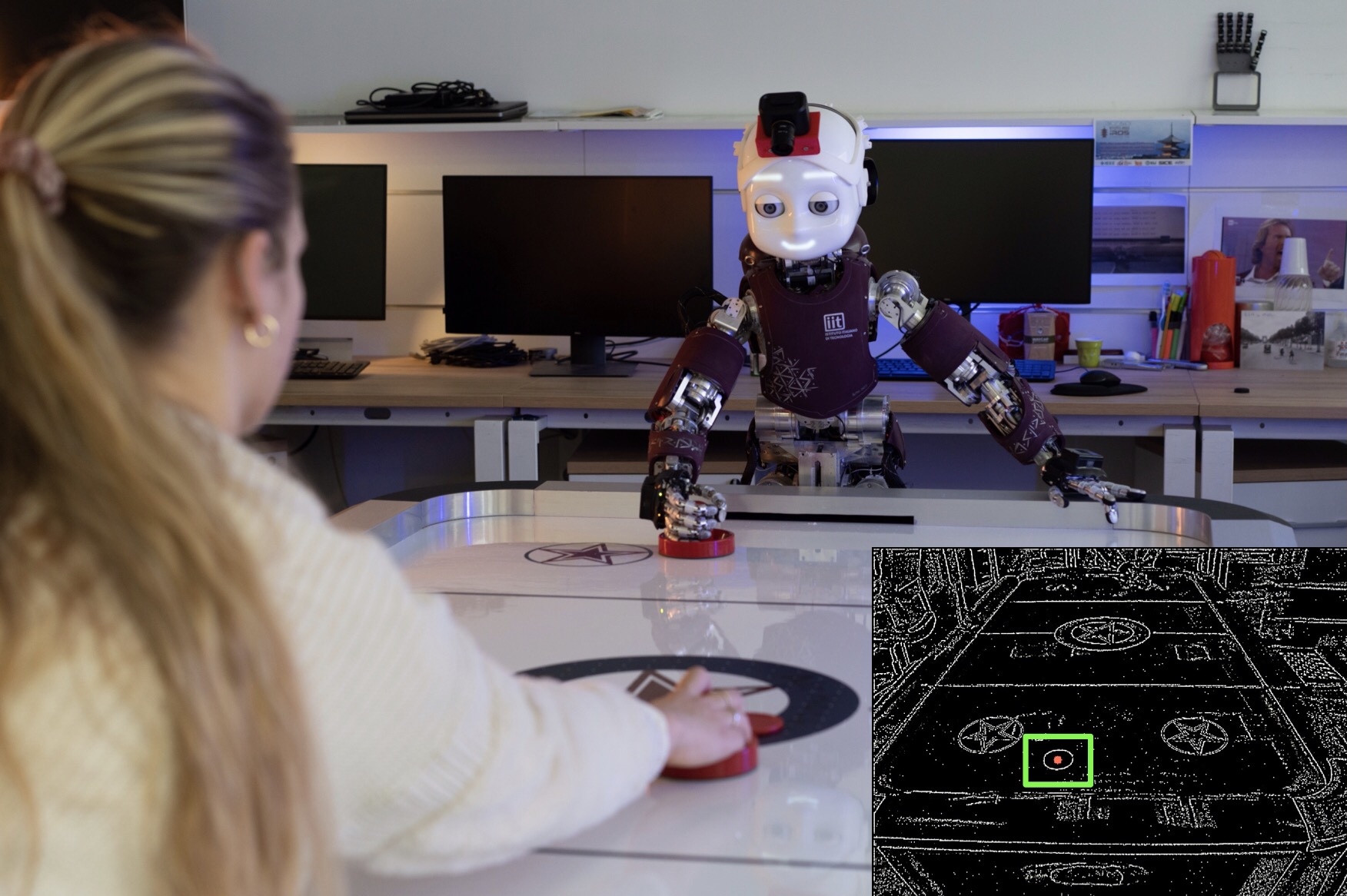}
    \caption{To investigate vision-action control loops with event-cameras we evaluate visual tracking algorithms applied to the task of playing air hockey. The game consists of two players hitting a round disk (a puck) with the paddle on a flat surface. The neuromorphic iCub~\cite{bartolozzi2011embedded} exploits event-based cameras for low-latency vision.}
    \label{fig:setup}
\end{figure}

\begin{figure*}
\centering
\includegraphics[width=\linewidth]{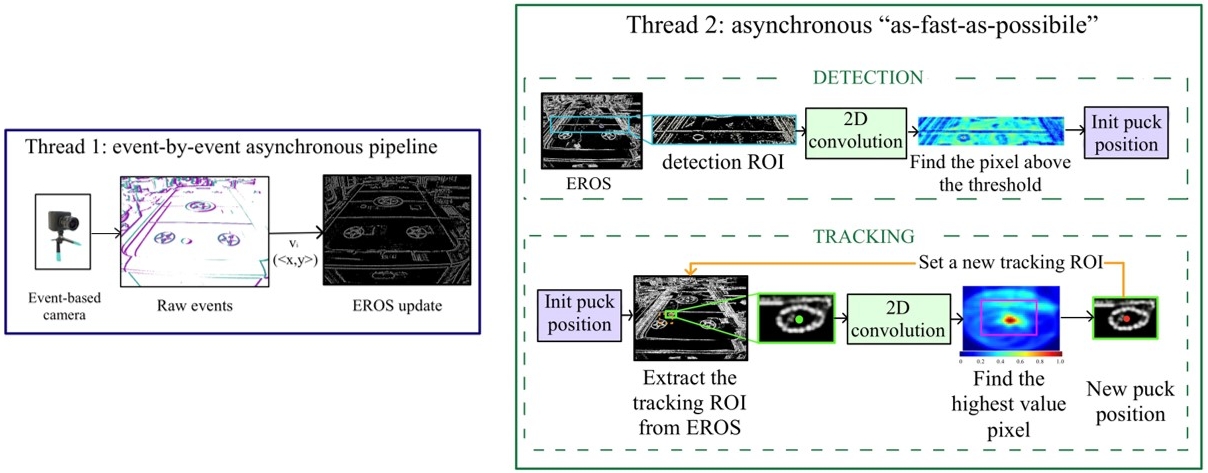}
\caption{Pipeline to track the air hockey puck using an event-based camera. Thread 1 is asynchronous and performed event-by-event. Thread 2 runs in parallel leveraging the event-based representation updated by Thread 1. Thread 2 performs the detection to initialize the puck position and it keeps computing the puck position.}
\label{fig:pipeline}
\end{figure*}

Camera technologies play a crucial role in system performances and recent years have seen a surge of event-cameras~\cite{gallego2020event} for robotics applications. They exhibit extremely low latency (in the order of micro-second), which makes them an excellent candidate for robotic applications with vision in the loop. In addition, event-cameras transmit only illumination changes at the time they occur and filter automatically redundant data: power and computational time are consumed to process only active pixels. An energy-friendly sensor is crucial for on-board computers or devices, towards the development of more autonomous systems. 

The key visual problem for air hockey is that of tracking the puck, and various tracking algorithms have been proposed to exploit event-cameras. Clustering-based methods were initially proposed for stationary cameras~\cite{litzenberger2006embedded,pikatkowska2012spatiotemporal,delbruck2013robotic,lagorce2014asynchronous,camunas2017event,zhu2017event} taking advantage of the sparse data coming from the camera. Their principle consists in processing each incoming event to determine whether it belongs to a cluster, for instance, based on a distance criterion~\cite{litzenberger2006embedded, barranco2018real} (inspired by mean-shift tracking) or assign features defined with Hough transform based on the Iterative Closest Point (ICP) principle~\cite{lagorce2014asynchronous}. However, for moving cameras, the assumption that all nearby events belong to the same object is no longer valid, as the relative movement between the camera and the environment will generate events from stationary items. As such, the tracking position can drift towards events coming from the background and other objects, eventually failing.

Tracking an externally moving object from a moving camera has been shown to be possible for a strong fiducial by means of a robust particle filter-based algorithm~\cite{glover2017robust}, for pattern-based tracking ~\cite{ni2015visual} and for long-term tracking using a local sliding window approach~\cite{ramesh2020tld}.

Finally, motion-segmentation algorithms have also been proposed to segment targets moving differently than the camera motion itself~\cite{mitrokhin2018event, falanga2020dynamic, he2021fast}. The algorithms perform best with constant camera motion, and with fast-moving, large targets. Given the iterative algorithm is typically a heavy operation, operating on large batches of events, and designed to solve a general task of motion segmentation, we opt to focus on tracking algorithms in which the constraints of the air-hockey game can be applied to achieve the lowest possible latency exploiting event-cameras.
In this paper, we investigate event-based tracking for the task of air-hockey, with the focus on investigating its accuracy and real-time latency, towards closed-loop control. To this end, we propose a novel tracking method that preserves the event-camera asynchronicity and maximises the event-throughput, in order to minimise computational latency, inspired by~\cite{glover2021luvharris}. In particular, we exploit the paradigm of simultaneous parallel processing of (i) the spatial surface representation, that is updated asynchronously, and (ii) the heavier tracking algorithm, based on convolution kernels that run on the most up-to-date surface representation as fast as possible, to detect and track the air hockey puck.

Experiments are performed using the neuromorphic iCub robot~\cite{bartolozzi2011embedded}, equipped with an ATIS~\cite{posch2010qvga} camera. This paper is focused on the characterisation of the visual pipeline and its robustness to the visual clutter produced by the movements of the robot and to the increase in computational latency due to the corresponding increase of data to be processed. 
Specifically, we first characterise the accuracy, latency, and robustness of the tracking with a stationary camera, and then moving the robot head simulating air-hockey related actions of the iCub, with head stabilisation.
We compare our results in terms of latency and accuracy to other event-based tracking algorithms~\cite{lagorce2014asynchronous}~\cite{glover2017robust}, by running these on the same two datasets.

\section{THE PUCK-TRACK ALGORITHM}

The pipeline of our system (named PUCK-track) for tracking an air hockey puck comprises the following components, which are integrated as illustrated in Fig.~\ref{fig:pipeline}:
\begin{itemize}
    \item The ATIS~\cite{posch2010qvga} event camera mounted on the iCub head;
    \item the Exponential Reduced Ordinal Surface (EROS) representation update (inspired by~\cite{glover2021luvharris});
    \item a detector that initializes the puck position at the beginning of the game;
    \item a Region of Interest (ROI) that follows the target and that is used to limit the area of the field of view over which the convolutions for the tracking are performed;
    \item a tracker that follows the puck once detected, while is moving on the table.
\end{itemize}

\subsection{Event-camera}
Events are transmitted from the camera in the form of a tuple $v=<x,y,t,p>$, where $v_x$ and $v_y$ are the pixel coordinates, $v_t$ is the time stamp and $v_p$ represents the polarity, which indicates the increase or decrease in light intensity. Our algorithm works only with $<v_x,v_y>$ information and the implicit order of pixels.


\subsection{EROS}
We propose an Exponential Reduced Ordinal Surface (EROS) as a representation of the visual signal coming from the camera, that is quickly updated for each and every event and subsequently used for detection and tracking. The EROS sits in-between the asynchronous, high temporal resolution event-camera and the slower processing algorithm. It is a key component that enables the decoupling of these processes to allow low-latency, high event-throughput, for complex visual processing algorithms. The EROS is similar to the Threshold Ordinal Surface (TOS)~\cite{glover2021luvharris} with a small modification to the representation decay method defined in Algorithm~\ref{alg:TOSalg}, where $k_{EROS}$ corresponds to the update region size around each event position. EROS can be used at any given point in time by downstream processing, as a 'grey image', with values between 0 and 255. Similar to the TOS~\cite{glover2021luvharris}, the EROS is filtered with a  Gaussian blur before further processing. However, as we only process a ROI for both detection and tracking, we perform the blur only on the selected area. 

\subsection{Puck Observation and Kernel Parameterization}\label{sec:kernel}

\begin{figure}
    \centering
    \includegraphics[width=\linewidth]{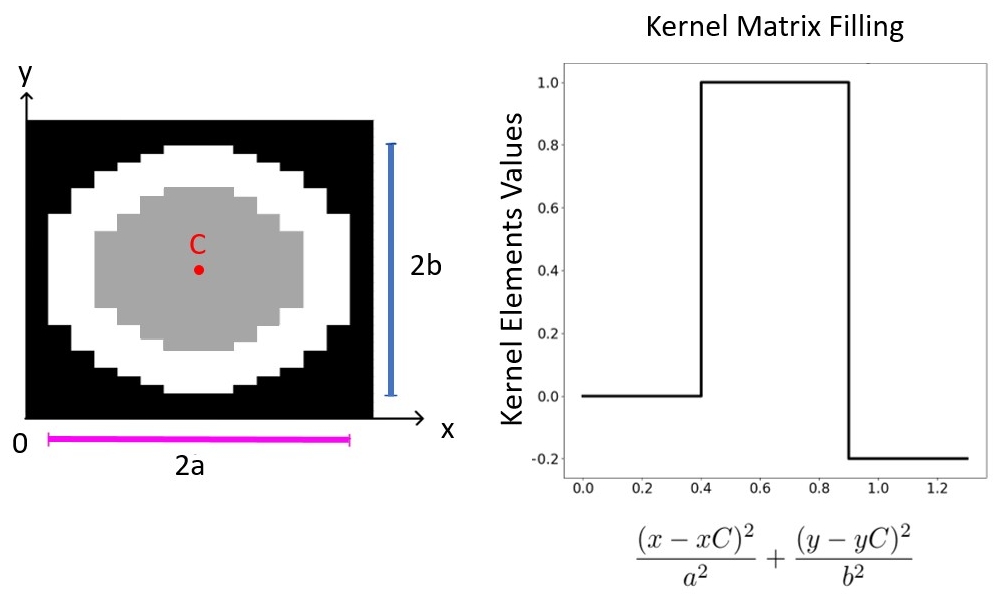}
    \caption{Kernel matrix construction according to ellipse parameters: center C, major and minor axes a,b. The image contains ones along the ellipse, zeros inside the ellipse and a tunable negative value outside the ellipse.}
    \label{fig:kernel}
\end{figure}

In our application, the object of interest is the air hockey puck, which is a round disc made of plastic. To detect and track the puck position, we run a two-dimensional convolution over the EROS representation, using a suitable kernel.
As shown in Fig.~\ref{fig:kernel}, the kernel is a matrix of weights built from the ellipsis equation:
\begin{equation}
    \frac{(x-xC)^2}{a^2}+\frac{(y-yC)^2}{b^2}=1
\end{equation}
The target projection on the image plane changes shape (e.g. width and height) as it moves forward and backward along the table. It can be a circle or an ellipsis of varying size according to the 3D puck position with respect to the camera frame.
\begin{algorithm}
\caption{Event-by-event EROS update}
\label{alg:TOSalg}
\begin{algorithmic}
\REQUIRE $d = 0.3^{1.0/k_{EROS}}, k_{EROS}$
\STATE \verb|for| $x = v_{x}-k_{EROS} : v_{x}+k_{EROS}$
\STATE \quad\verb|for| $y = v_{y}-k_{EROS} : v_{y}+k_{EROS}$
\STATE \qquad$EROS_{xy} \gets EROS_{xy} \times d$
\STATE $EROS_{v_xv_y} \gets 255$
\end{algorithmic}
\end{algorithm}
This implies that the kernel needs to change consequently. Assuming that the object moves on a plane and that the robot distance from the table is fixed, we can determine the parameters $a$ and $b$ for any given puck position on the image plane. To this end, we run an offline calibration with the robot in place. A function $f$ which relates the puck coordinates to its estimated width and height was obtained via multi-linear regression from a set of manually annotated puck observations $x_i,y_i,a_i,b_i$:
\begin{equation}\label{eq:kernel_size}
  \begin{array}{l}
    a_i = f(x_i, y_i) =  k_0 + k_1*x_i + k_2*y_i\\
    b_i = f(x_i, y_i) = h_0 + h_1*x_i + h_2*y_i
  \end{array}
\end{equation}

where $x_i$ and $y_i$ are the puck coordinates, $a_i$ and $b_i$ are respectively the puck width and height, and $k$ and $h$ are the resulting coefficients of the regression.
Exploiting $f$ we can precompute several kernels combining different widths and heights when initializing the algorithm. This allows selecting the closest shape to the current object projection, without recomputing it at every iteration.

\subsection{Parallel Implementation}

A key feature of the PUCK-track algorithm is the decoupling of the event-by-event processing from the relatively heavier computation of the detection of the puck position, as proposed in~\cite{glover2021luvharris}. As the camera potentially produces $> 10$ million events per second, any complex processing performed event-by-event does not meet the real-time constraint in highly dynamic robotic tasks. This worsens for cameras that move with the robot and for cluttered scenarios. To achieve real-time event-processing and ``as-fast-as-possible'' puck detection, two asynchronous threads run in parallel, and the computation of each one is decoupled from the other, as shown in Fig.~\ref{fig:pipeline}:
\begin{itemize}
    \item The first thread updates the EROS representation event-by-event. Since the update operation is light, this part is capable to handle a high event-throughput from the latest generation and high-resolution event-cameras.
    \item The second thread computes the puck position as-fast-as-possible on the most up-to-date EROS representation and it can send the puck position asynchronously for potential robot control. The heavier convolution and filtering operations are performed on a single image-like representation, therefore the computational time does not depend on the event-rate.  
\end{itemize}

\begin{figure}
    \centering
    \includegraphics[width=\linewidth]{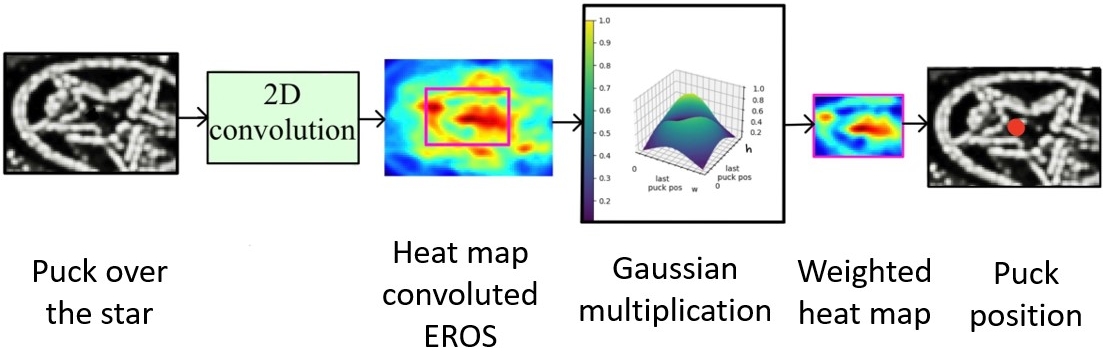}
    \caption{Gaussian multiplication to deal with multiple features inside the ROI (e.g. when the air hockey puck passes over a star shape printed on the table surface.}
    \label{fig:gaussian_mul}
\end{figure}

\subsection{Detection}

Detection occurs using the parametrized kernel described in Section~\ref{sec:kernel}.
The detection takes place inside a predefined area in the middle of the table that is free from distracting shapes, highlighted with a blue rectangle in Fig.~\ref{fig:pipeline}:
\begin{alignat*}{2}
 ROI_d = <x_d, y_d, w_d, h_d>
\end{alignat*}
where $(x_d,y_d)$ is the top-left corner and $(w_d,h_d)$ correspond respectively to the rectangle width and height. During this phase the kernel has fixed dimensions given by $f(x_c, y_c)$ (Equation \ref{eq:kernel_size}), being $x_c$ and $y_c$ the coordinates at the center of $ROI_d$.
Once the puck is detected, the tracking region of interest $ROI_t$ is initialized centered in the detected position and with dimensions slightly bigger than the projected puck width and height.

\begin{figure*}
    \centering
    \includegraphics[width=\textwidth]{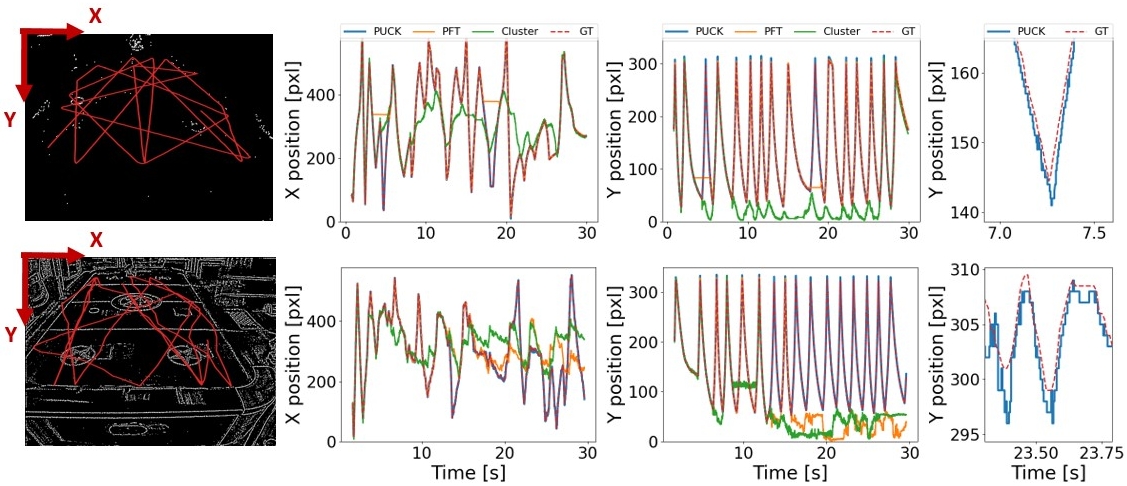}
    \caption{Qualitative comparison of tracking algorithms: our method (PUCK), Clustering-based Tracker (Cluster) ~\cite{lagorce2014asynchronous} and the Particle Filter Tracker (PFT)~\cite{glover2017robust}. One sequence for each experiment is taken as an example: static (first row) and moving camera (second row). The left column shows the trajectory of the puck over time, superimposed on the EROS representation. The difference between static and moving cameras can be appreciated by the presence of the background decorations of the air-hockey table in the moving scenario. The central and right columns show the actual $x$ and $y$ position of the puck (GT) and their estimate over time for the three tracking methods. PUCK and the ground-truth overlap for the full duration of the dataset. The right column shows more in detail the PUCK tracked $y$ position compared to the ground truth.}
    \label{fig:x_y_position}
\end{figure*}

\subsection{Tracking}

During the tracking stage, the convolution is done inside the $ROI_t$, highlighted in green in Fig.~\ref{fig:pipeline}:
\begin{alignat*}{2}
 ROI_t = <x_t, y_t, w_t, h_t>
\end{alignat*}
where $(x_t,y_t)$ is the top-left corner computed from the latest detected puck position, whereas the width and height $(w_t,h_t)$ are slightly bigger than the puck size. Setting a restricted ROI around the puck is efficient because computational power is not wasted doing calculations with events not belonging to the object of interest. The output of the convolution is used to find the highest peak corresponding to the puck position. As the object moves, the tracker updates the position and dimensions of $ROI_t$ and then continues to perform convolutions on the latest EROS updated with new events.

To improve the robustness of the puck observation, we multiply the result of the convolution with a two-dimensional Gaussian kernel centered in the last tracked position and with a standard deviation equal to the $50\%$ of the kernel width and height. In this way, we give more priority to the locations closer to the previously known puck location. Moreover, the maximum is computed only inside a restricted region around the last tracked position, smaller than the previously mentioned region of interest (highlighted in magenta inside the heat map in Fig.~\ref{fig:gaussian_mul}). 

\section{EXPERIMENTS AND RESULTS}

We evaluated the accuracy and latency of the proposed visual tracking algorithms for the task of tracking the air-hockey puck. The reaction-time and stability of the robot when in closed-loop control will be dependent on both these metrics. The proposed PUCK-track algorithm was compared to an event-based cluster tracking algorithm (Cluster)~\cite{lagorce2014asynchronous} and an event-based Particle Filter Tracker (PFT)~\cite{glover2017robust}.

\subsection{Comparison Algorithms}

Cluster is chosen as a light-weight event-by-event processing algorithm. Each new event that falls too far from the puck position is ignored; all other events are used to update the puck position and shape as described in~\cite{lagorce2014asynchronous}. The update requires only a few CPU operations, but the algorithm assumes all nearby events are part of the object to be tracked and therefore the algorithm is not discriminative towards the puck itself.

PFT~\cite{glover2017robust} was designed to robustly follow a single provided visual target, even in cluttered conditions where many events are generated from the relative motion between the camera and the background. The algorithm dynamically estimates the optimal temporal window for the observation of the target along with the pattern positions and size. However, as multiple events need to be processed for each particle, for each filter update the algorithm is more computationally heavy. The algorithm was modified to track an ellipse shape instead of a circle to better fit the air-hockey data. Specifically, in addition to the $x$ and $y$ target position, it estimates the ellipsis width and height, but keeps the ratio between them fixed over time.

Cluster and PFT do not have an inherent detector, therefore to compare each algorithm fairly these algorithms were initialised at the same positions as calculated by the PUCK-track detection operation. Accuracy is calculated as the distance between the tracked positions and the manually-labeled ground truth. Latency is measured at the end of each algorithm update, by computing the temporal period of events that have not yet been processed by the algorithm - thereby measuring the time difference between the point the algorithm has processed, and the true time passed. 

\subsection{Hardware}
Experiments were performed with an ATIS HVGA Gen3 (640x480 pixels) mounted on the neuromorphic iCub~\cite{bartolozzi2011embedded} head. The air-hockey table measures 213.5\,$\times$122\,cm, and its height is 81\,cm. There are several ``star patterns'' and lines that produce contrast (hence events from the moving camera). The puck bounces off the table edges with little loss of velocity. The air hockey puck measures 6.5\,cm. The robot is positioned 68\,cm far from one end of the air hockey table to have the whole playing field inside the visual space. Since the human opponent hits the puck using the paddle from the other end of the table (see Fig~\ref{fig:setup}), the motion of the opponent is also visible in the camera, and the hand motion and puck motion coincide at each puck hit, providing an even more complex scenario for the tracking. All code was implemented in C++ and run on a PC (Intel i7 @ 2.60GHz with 16 GB of RAM).

\subsection{Datasets}
The algorithms were benchmarked in two different scenarios at increasing complexity: in the former, the robot does not move and the camera is \textit{static}, in the latter, the robot moves as if playing a real air-hockey game. Despite head stabilisation, also the camera moves and generates additional events from the table and the surrounding scene, as shown in Fig.~\ref{fig:x_y_position}. Both scenarios comprise 10 sequences of 30\,s each. We compute the $x$ and $y$ coordinates of the puck position with precise timing information over the length of all 20 sequences, and the comparison between the three algorithms is performed with respect to the manually annotated ground-truth (GT).
The \textit{moving} scenario is performed to replicate a similar type of motion expected during real play, the end-effector is constrained to move parallel to, and the full length of, the air hockey goal in a defensive manner, while the head is also moved to compensate the torso motion, to keep the camera as still and straight as possible. The end-effector was moved with a mean speed of 30\,cm/s, which leads to head yaw oscillations of approximately 6 degrees, still more than sufficient for the camera to respond to contrast change across the full scene.

\subsection{Results: Accuracy}

Across all datasets, PUCK-track successfully tracks the puck for the entire duration of the sequence, as qualitatively shown by the PUCK and ground-truth trajectory overlap in Fig.~\ref{fig:x_y_position}. 

PFT tracks for the majority of the dataset for the \textit{static} example, however does lose track for short periods. PFT also tracks in the \textit{moving} case but fails for significant portions, as discussed in the next section.

Cluster begins tracking the puck, however loses track at the first collision of the puck with an external object that produces events. In the case of the \textit{static} camera, it is typically the moving hand of the opponent player, in the case of the \textit{moving} camera it is as soon as the puck passes ``star'' patterns drawn on the air-hockey table.

\begin{figure}
\begin{subfigure}{.5\textwidth}
  \centering
  \includegraphics[width=.8\linewidth]{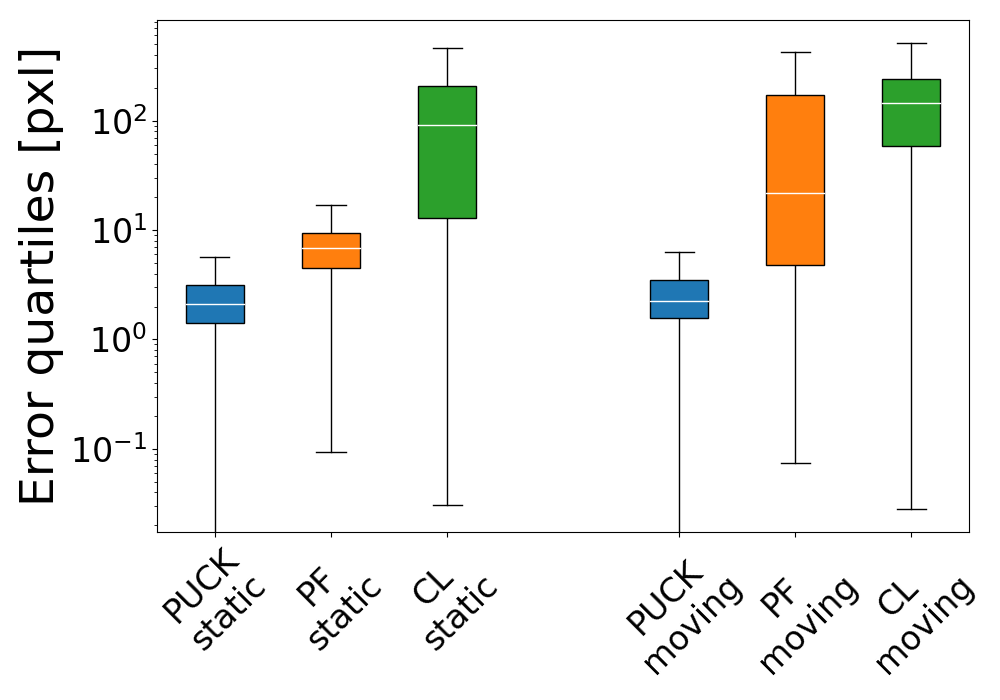}
  \label{fig:mean_error}
\end{subfigure}
\begin{subfigure}{.5\textwidth}
  \centering
  \includegraphics[width=.8\linewidth]{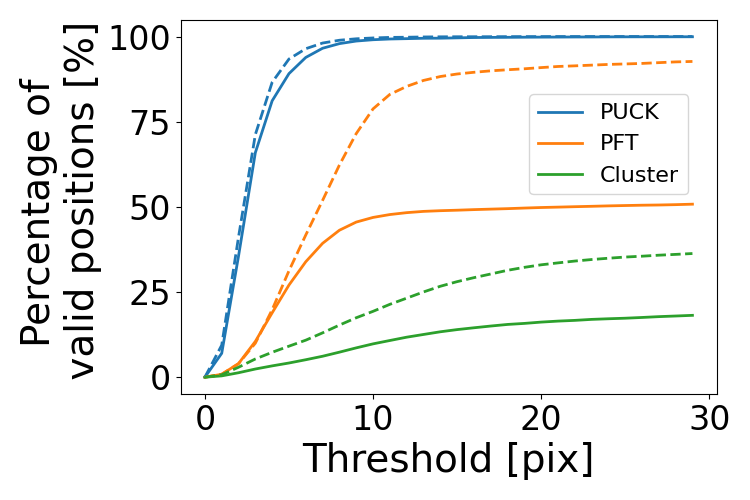}
  \label{fig:valid_pos}
\end{subfigure}
\caption{Tracking accuracy results for 10 sequences and both experiments, comparing our method (PUCK) to Particle Filter Tracker (PFT) and Cluster algorithms. (a) The box plot shows the median, quartiles and extreme values in logarithmic scale. (b) The percentage of valid positions is marked with dashes for the static camera, while for the moving camera is represented by straight lines.}
\label{fig:accuracy_summary}
\end{figure}

The median error over all datasets of the PUCK-track algorithm is an order of magnitude smaller than the PFT and almost two orders of magnitude smaller than the Cluster, as shown in Fig.~\ref{fig:accuracy_summary}. The mean error of PUCK-track resulted to be about 2.47 pixels (static camera) and 2.81 pixels (moving camera). The error is consistent for PUCK-track between \textit{static} and \textit{moving} scenarios indicating its robustness to reject non-puck-like objects. The variance of the PFT increases for the \textit{moving} datasets as tracking becomes less consistent under more difficult conditions. Cluster has a higher median, but lower variance, for the \textit{moving} datasets, indicating it is consistently always in failure. 

PUCK-track tracks more closely to the ground-truth position, compared to PFT as indicated by the lower pixel threshold achieved in Fig.~\ref{fig:accuracy_summary}. PFT estimates the temporal window of events required to detect the puck position, however, if it is slightly incorrectly estimated, it can result in inaccurately placing the exact centre of the target, even if tracking is maintained. Using a threshold of 3.5 pixels (static camera) and 4 pixels (moving camera) results in $80\%$ of the positions for all sequences to be detected correctly.


\subsection{Results: Latency}
To better estimate the true latency of the algorithms, we performed a live tracking experiment, in which the robot was moving as the opponent continuously hit the puck. Conditions were consistent for each algorithm, however, they were run individually so as not to interfere with each other. The PFT had the highest mean latency of approximately 4\,ms, as it has the highest computational load, and each event may be processed more than once. PUCK-track achieves a mean latency on the order of 8\,$\mu$s, while Cluster achieves 4\,$\mu$s.

The PFT produces an output on the order of 200\,Hz, PUCK-track produces an output typically over 1\,kHz, and Cluster performs an update for each event within the nearby region, possibly over 1\,MHz.

\begin{figure}
\centering
\includegraphics[width=0.9\linewidth]{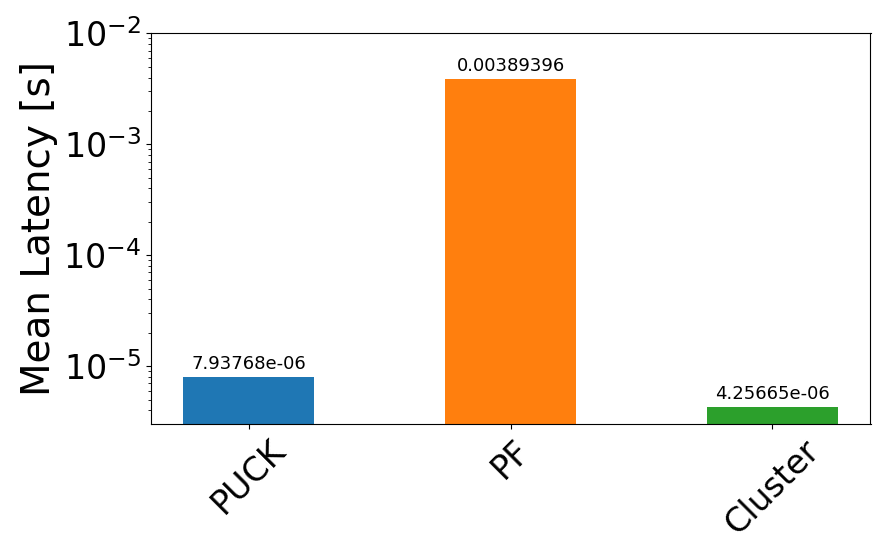}
\caption{Mean latency for the three event-based tracking algorithms, including our method, measured during an online experiment and represented using the logarithmic scale.}
\label{fig:mean_latency}
\end{figure}









\section{DISCUSSION}

The Cluster algorithm is a traditional ``event-based'' algorithm in that a small update is performed for each event produced by the camera. It can achieve the lowest latency as any event that is not close to the known target position can be discarded without processing. However, as the algorithm is not discriminative to the puck shape, the algorithm itself is not suitable for the tracking application required by the air-hockey task (even for a static camera). It has also been shown that event-by-event algorithms don't scale for complex operations with high-resolution, fast moving cameras~\cite{glover2021luvharris}. The parallel ``as-fast-as-possible'' approach we followed from~\cite{glover2021luvharris} instead keeps the event-by-event throughput extremely high, while also allowing for complex visual processing to occur. It was not tested if PUCK-track could still achieve a high accuracy if the EROS was calculated only within the target ROI, doing so would further reduce the latency of PUCK-track.

The PFT was designed to reject background events and robustly follow a target~\cite{glover2017robust}, however, in our experiments, the tracker did fail at some point on all datasets. We believe that the PFT could achieve higher accuracy by tuning the algorithm better for the task, and applying task-specific constraints. However, even a better tuned and modified PFT would produce a higher latency than PUCK-track, hence we deemed the work to improve PFT excessive. The following algorithmic differences also explain the performance gap between PFT and PUCK-track:
\begin{itemize}
    \item The PFT visually estimates the puck size, similarly to the best fit. Such an approach is flexible but more prone to false-positive detections. It was found that constraining the visual size based on $x$ and $y$ target position since the object moves on a plane reduced tracking errors for PUCK-track. 
    \item The PFT dynamically estimates the batch of events to use such that a full ellipse is visible. Failure can occur if the value is not appropriate. The EROS performs the same role for PUCK-track and, for this task, achieves a more consistent visual representation.   
    \item The PFT is a stochastic process, allowing a fixed number of samples (i.e. particles), and hence bounds computation independently of the state dimensions. PUCK-track uses convolutions, which results in a dense sampling, which scales up with state size and dimensionality.
\end{itemize}
A PFT could be created that used the EROS and applied task-specific constraints for air-hockey, leaving the only difference being whether to use sampled (particle filter) or dense (convolutions) measurement of puck positions. In 2D space, with a small ROI, the random sampling of PFT could introduce more problems than are solved.

\section{CONCLUSIONS}

The proposed PUCK-Track algorithm reached a better compromise between tracking accuracy and low computational latency compared to previous event-based algorithms for object tracking. Latency in the order of $\mu$s was obtained using the parallel ``as-fast-as-possible'' method introduced in~\cite{glover2021luvharris}. The accuracy of PUCK-track also improved on state-of-the-art algorithms designed for such moving camera conditions. The EROS provided a robust visual representation on which target observations could be made with high consistency and precision. However, even to solve a seemingly simple task, certain task-specific constraints had to be imposed, such as 2D motion and the object size.


With this work, we add evidence that event-cameras can be used for low-latency, high-frequency tracking, which is a key component for highly reactive, but stable, vision-in-the-loop robot control. Such processing can be achieved on standard CPU, with expected computational advantages over high-frame-rate traditional previously shown to be required for a task as dynamic as air-hockey.

Vision-based control applications should not make strong simplifying assumptions about the environment (e.g. static scenarios), or suppose a previous knowledge of the environment. However, designing a specific application, as the air hockey game, can help to better understand challenges behind general real-world scenarios and exhibit different dynamics.

\addtolength{\textheight}{-12cm}   







\bibliographystyle{IEEEtran}
\bibliography{IEEEabrv,IEEEexample}

\end{document}